# Representing scenarios for process evolution management

*2018, Anton Kolonin, Aigents Group ([http://aigents.com](http://aigents.com)), SingularityNET ([https://singularitynet.io/](https://singularitynet.io/))*

**Abstract:** In the following writing we discuss a conceptual framework for representing events and scenarios from the perspective of a novel form of causal analysis. This causal analysis is applied to the events and scenarios so as to determine measures that could be used to manage the development of the processes that they are a part of in real time. An overall terminological framework and entity-relationship model are suggested along with a specification of the functional sets involved in both reasoning and analytics. The model is considered to be a specific case of the generic problem of finding sequential series in disparate data. The specific inference and reasoning processes are identified for future implementation.

## 1. Introduction

Initially, the scope of this work was bound to the problem of detection and prevention of undesirable scenarios of human behavior such as conflicts of interests, based on data found in online media sources and open data. Starting from the problem of event evolution study [1], we have attempted to derive a unified terminological system and data model capable of performing causal inferences on processes found in different kinds of data so the predictions can be made, undesired developments can be detected early and preventive measures can be taken in a timely fashion. Since it was assumed that the source data may be extracted from text sources, this study was bound to textual data, primarily. Respectively, we were assuming to use Aigents platform ([http://aigents.com](http://aigents.com)) with its underlying ontology [2] for event detection in online texts [3] so the design and terminology are based on Aigents ontology terms. During our study, we have derived a data model and have identified functional sets that may be applied to solving different sorts of problems involving causal and temporal inference on different series of data in different domains.

## 2. Definitions and goals

Two fundamental concepts underlying this in Aigents platform [2] are the "**thing**" (may be called "**entity**") and the "**set**" (may be called "**collection**").

**Thing (Entity)**
- may have one or more (**properties**);
- may be identified by one or more textual **patterns** (may be called **templates**) through a property called "**patterns**" [3];
- has a property called **name** which represents textual pattern by default implicitly;
- may be an instance of a thing of a higher abstraction level (the latter may be thought as a **class** in such case).

**Set (Collection)**
- may have one or more things or other sets as members;
- may be disjunctive, including logically sufficient members – representing alternatives in style of disjunctive logical operation OR (called Any-Set), in textual notation framed with braces "{...}";
- may be conjunctive, including logically necessary members – representing all conditions in style of conjunctional logical operation AND (called And-Set), in textual notation framed with parentheses "(...)";
- may be conjunctively-ordered, including necessary members in required order – representing all conditions in style of conjunctional logical operation AND with necessary order of following (called Seq-Set), in textual notation framed with brackets "[...]".

For analysis of sequential chains of events and causal inference on these chains, the following categories of things may be considered.
- **Actor** — a thing which has all its properties and patterns identifying the specific actor in some relationship in some particular **role**, such as "water", "stone", "grinding", "management", "John Doe", "HealthPlus Ltd.", "US", "UN".
- **Role** — a thing which identifies participation if and **actor** in some **appearance** or a role in which a given **actor** plays in given **appearance**.
- **Appearance** — a thing with one or more **actors** playing different **roles**, being properties of given appearance as a thing, such as in two examples: "the water grinds the stone" ("water" in the role of "subject", "grinds" in the role of "verb", "stone" in the role of "object"), "John Doe leads the HealthPlus Ltd." ("John Doe" in the role of "director", "management" in the role of "affiliation", "HealthPlus Ltd." in the role of "company").
- **Event** — an **appearance** possessing property of time, such as moment of time, time period or set of times or time periods (called **times**).

Deeper analysis of the appearances and events representing them may lead to consideration of the following sets of things.
- **Connected Appearances** – conjunctive set of appearances having the same actor in the same (if **specifically connected**) or different (if **generally connected**) roles, such as in the examples: "the water grinds the stone", "the water is source of line", "the water turns into steam".
- **Tightly Connected Appearances** – conjunctive set of appearances having more than one of the same actors in the same (if **specifically tightly connected**) or different (**if generally tightly connected**) roles, such as in the examples: "John Doe leads the Healthcare Department", "John Doe leads the Welfare Fund", "John Doe leads the HealthPlus Ltd.".
- **Coincidence** – conjunctive set of appearances sharing the same "**times**" property (connected by time) or simultaneous events in other words.
- **Connected Coincidence** – conjunctive set of appearances or events sharing not only the "**times**" property, but some actor.
- **Tightly Connected Coincidence** – conjunctive set of appearances or events sharing not only the "times" property, but more than one actor.
- **Scenario** – conjunctively-ordered set of appearances or connected appearances with order implying temporal sequence, which may be implemented one or more times in different **processes**.
- **Process** – conjunctively-ordered set of events, coincidences or connected coincidences implementing particular scenarios.

All of the things, appearances and scenarios may be of different degrees of abstraction, such as the following:
- Thing "Jonh Doe" may have different instantiations in different appearances and events such as "John", "Johnny", "Mr. John Doe", "SSN#1234-56-7890" etc.;
- Appearance "John Doe leads the HealthPlus Ltd." may be instantiated in more than one way, such as ""HealthPlus Ltd. has John Doe as a chief", "Chief of HealthPlus is that guy", "Her husband is CEO of HealthPlus Ltd.", etc.
- Scenario "John Doe headed a budgetary institution, after which he has founded a commercial company" may be instantiated "After when Mr. Doe was appointed to lead the Healthcare Department, his wife has founded commercial company HeartyCare Ltd." or "John's wife has founded HealthyCare Ltd. soon after her husband entered management board of Healthcare Department".

Based on scenarios, composed of appearances, and processes, composed of events, more complex sets may be derived for further processing, as follows.

- **Alternative Actor** – a disjunctive set of actors that may play a specific role in an appearance with high degree of abstraction, and may be treated as domain of admissible values for a role. For instance, sets X = {"John Doe", "Jane Roe"}, Y = { "founded", "established", "registered" } and Z = {"HealthPlus Ltd", "HeartyCare Ltd.", "HealthyCare Ltd."} may be used as actors of the highly abstract appearance "X Y Z" with "Jane Roe founded HealthyCare Ltd." as one of possible instances.
- **Situation** (may be called **Context**) – set of appearances, being not necessarily connected to each other, yet being abstract classes in respect to a substantial number of coincidences, that is – being a specific instance of the generic case of some simultaneously happening appearances. One example is a conjunctive set ("X Y Z", "A is wife of X", "A B C"), where B is disjunction {"lead", "head", "manage"}, C is disjunction {"Healthcare Department", "Department of Health"}, Y is {"founded", "established", "registered"} and Z is disjunction {"HealthPlus Ltd", "HeartyCare Ltd.", "HealthyCare Ltd."}, so we represent a situation where someone heads a governmental healthcare department while her husband has founded a business company in the healthcare domain, which may potentially represent a corruption case.
- **Situational/Contextual Fork** – a disjunctive set of appearances or situations which have nearly equal probabilities of taking place in the course of the development of an abstract scenario or a specific process.
- **Situational/Contextual Trigger** – an appearance or an event, which when added to a situation preceding a situational fork may complement the latter by changing the probabilities of its alternatives, so that the appearance or event may implicitly or explicitly, depending on distribution of probabilities, change its development to a process to follow another scenario.

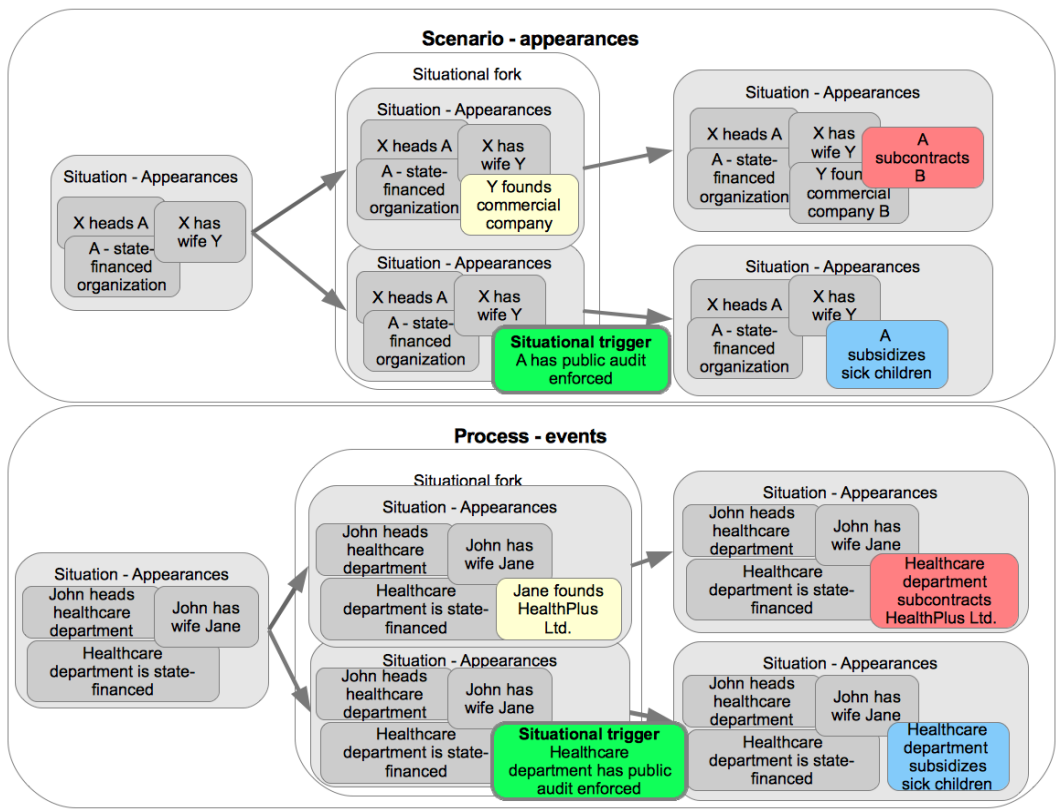

**Fig. 1.** The top half of the figure represents an example of an abstract scenario, which includes a sequence of situations not attached to any particular moment of time, involving appearances with abstract actors. The bottom half of the figure represents a specific process instantiating the scenario above with real actors performing in real time. In both cases, there are abstract (at the top) and specific (at the bottom) situational forks and triggers that enable the development of the scenario or process to head in one direction or another.

For the causal analysis within the conceptual framework described above, the following tasks may be considered:
identification of events and actors participating in them – based on patterns/templates representing them in the textual sources such online ads and social media on the Internet;
- determination of processes and their constituent coincidences, events and actors based on the scope of identified events;
- formation of the scope of possible abstract scenarios and determination of typical situations, appearances and sets of possible actors involved in the scenarios – based on the scope of determined processes;
- identification of overlapping and alternative scenarios and situational forks at their intersections – on the basis of all formed scope of scenarios;
- determination of situational triggers, which may effect the development of particular scenarios when executed with specific processes upon reaching one or another situational fork – based on the scope of identified situational forks.

## 3. Identification of events and actors by means of things and patterns.

As the main source of identification of events and their actors playing roles in different appearances could be online information on the online pages and in social media sources, we consider it necessary to extract the information from the web, folowing the earlier works [4,5]. For that purpose, let us see how diverse world of "things" and patterns representing these things may be described within an ontological model with the pattern description language of the Aigents platform [2,3].

**Actor** is considered a named thing or a thing possessing a property called **name**. To identify the actor thing in the text, template maybe taken implicitly from the name of the property itself, or from the designated property called patterns, which can supply more than one alternative pattern for a thing. On itself, each of the alternative patterns may be any set of textual strings and sets (And-Set, Or-Ser or – Seq-Set) enclosing other textual strings and sets recursively. Below are the examples how the named entities and their patterns are defined in Aigents, with all of them having the same effect, given simplicity of the example.
- Implicit definition of the pattern via thing name as disjunctive string set: *There name "{'john doe' 'jane roe'}"*.
- Explicit definition of the pattern via patterns property as disjunctive string set: *There name person. Name person patterns "{'john doe' 'jane roe'}"*.
- Explicit definition of the pattern via patterns property as multiple string patterns, associated as disjunction by Aigents engine behind the scene: *There name person. Name person patterns "john doe", "jane roe"*.

**Appearance** is a named or not named thing, which may have variables in its patterns, specified implicitly or explicitly. Each of the variables identify a particular property of the thing, corresponding to particular role that thing filling the property may play, being an actor in the appearance. Accordingy [3], variables are prefixed with dollar sign. For instance, the abstract **appearance** of someone buying something with pattern "$buyer buys $purchase" identifies two variables "$buyer" and "$purchase" which may be filled with specific actors corresponding to the roles, at the point a specific **event** is created to instantiate the **appearance**.
- When no variables are present in a pattern, the appearance might be though **nullary** (arity = 0) with an event instantiating the appearance corresponding to a single "anonymous" actor identical to the appearance itself, for example, thing with pattern "{'trump' 'us president'}" would correspond to the president itself as well as to a mention of the president in any context at the same time. Its occurrence in text will effect in creation of corresponding event.

- With one variable present in a pattern of an appearance, we might consider it **unary** (arity = 1), so the value of variable would correspond to a single actor involved in the context of the appearance, such as appearance with pattern "{'trump' 'us president'} {said told announced} $matter" would have subject of public saying made by US president as appearance actor called "subject".
- With more than one variable in an appearance's pattern, there could be corresponding named actors in the roles, identified by names of the variables (arity corresponding to number of variables), for example the pattern "{obama trump} {forced suggested} $organization to {impose implement apply} sanctions against $target" matching the text "Obama forced the EU to impose sanctions against Russia" would create event with two actors identified as organization=EU and target=Russia.

Patterns for appearances are identified in the same way as patterns for actors, for example, an appearance of someone saying something may be defined like this: *There name person_saying_something patterns "{John Jane Joe Joi} {said says told tells} $something"*.

**Event** is defined as an instance of some appearance class, and has this class filling its inheritance property '**is**". Also, it has a temporal property called "**times**" – filled with values corresponding to the time of the event discovery. The other two properties of event are filled in upon its creation are **sources**, filled in with a URL/URI with reference to the original text source, and **text** – attributed in with the text pattern extended so that variables are filled in with actual values matched in the text.

**Type system of thing properties and template patterns**

Since we have things and their properties implementing appearances and their roles, and the variables of the patterns are used as attributes of these properties-roles-variables when the thing-appearance-pattern is instantiated, let us consider some way to control the process of what such attribution by means of a type system that determines the domains of the values. In the present version of Aigents, by default, the presence of a variable in a pattern does not cause the creation of things-actors when things-events are instantiated, the textual references to the actors are just used to fill the "text" property of an event. However, there is a way to force instantiation of the appearance properties with specific matched values with an explicit definition of the appearance's properties, using the special property called "**has**". Moreover, there is a way to impose restrictions on the domains of the values to be used to fill the property and match the variable in the pattern. This can be done associating the variable-property-role with one of the pre-defined atomic types: **word, time, number, and money**. The latter association is accomplished by referencing role by name supplying it with an inheritance "**is**" property referencing one of the types, for example: "*There name item_quantity_cost patterns "On sale: $item, quantity $amount, prices $cost", has item amount, cost. Cost is money. Amount is number. Item is word.*"

In the forthcoming versions of Aigents, we plan to support the creation of complete events as appearances including actors attributed as appearances, not as text. To achieve that, the types of the properties and respective variables may be not atomic, but composite being built up as complete appearances with patterns and possibly other variables. It may look like this: "*There name person_doing_something patterns "$person $did $something", has person, did, something. Person is '{John Jane Joe Joi}'. Doing is '{{said told} {wrote printed}}'*".

# 4. Data model and mathematical model

According to the definitions given above, we introduce the data model represented on Fig.2, involving relationships of inheritance (**is**), possession (**has**), association with time (**times**) and inclusion of different sets — conjunctive (**and-set**) and conjunctively-ordered (**seq-set**). To introduce a mathematical model built upon the data model, let us start with some extra basic definitions. We will consider abstract things, not specifically bound to a particular time (roles, appearances, situations and scenarios as well as specific instantiations of these abstract things (actors, events, coincidences and processes, respectively), bound to specific moments or periods of times. It worth noticing that the boundary between abstract and specific may be blurred by adding more applicable moments of time to some specific thing until it gets associated with all possible moments of time and becomes completely abstract, however we will talk about abstract and specific things to keep further discussion simpler.

G – the set of all possible abstract and specific things (roles, actors, appearances, events, situations, coincidences, scenarios processes) present in given information space, assuming that given space may be represented with a graph in which case each particular thing would be vertex or node g in such graph.
R – the set of all possible roles or subset of G identifying the set of all things representing the roles involved in all the appearances in their relationships to other things as roles r, representing types of relationships between appearances and the other things.
A – the set of all possible actors or the subset of G identifying the domain of admissible values for instantiation of roles in all possible appearances, where each actor may be connected by an inheritance relationship with all of the roles that it plays.
E – the set of all possible appearances or subset of G identifying all possible combinations of roles that may be played by different actors involved in these appearances.
S – the set of all possible situations s as subset of G which enumerates all possible situations that may be experienced involving known appearances.
Q – the relative time identifying the order of following situations involved in scenarios as q.
O – the set of all possible scenarios o as subset of G including all known situations in the process of their following given the relative time identified above.
T – the time interval to consider, with point of time identified as t where t lies in range from 0 to infinity, with some increment along discrete temporal axis.

V – the set of all possible events or subset of G identifying every specific event v being appearance of particular appearance at given moment of time.
C – the set of all possible coincidences or the subset of G defined as all superpositions of all events present on the time axis, with each such superposition of events at the time t identified as c.
P – the set of all possible processes as the subset of G which includes all processes that had place or are having place in a given temporal continuum, where each process p includes a sequence of events with increasing values of the time property.

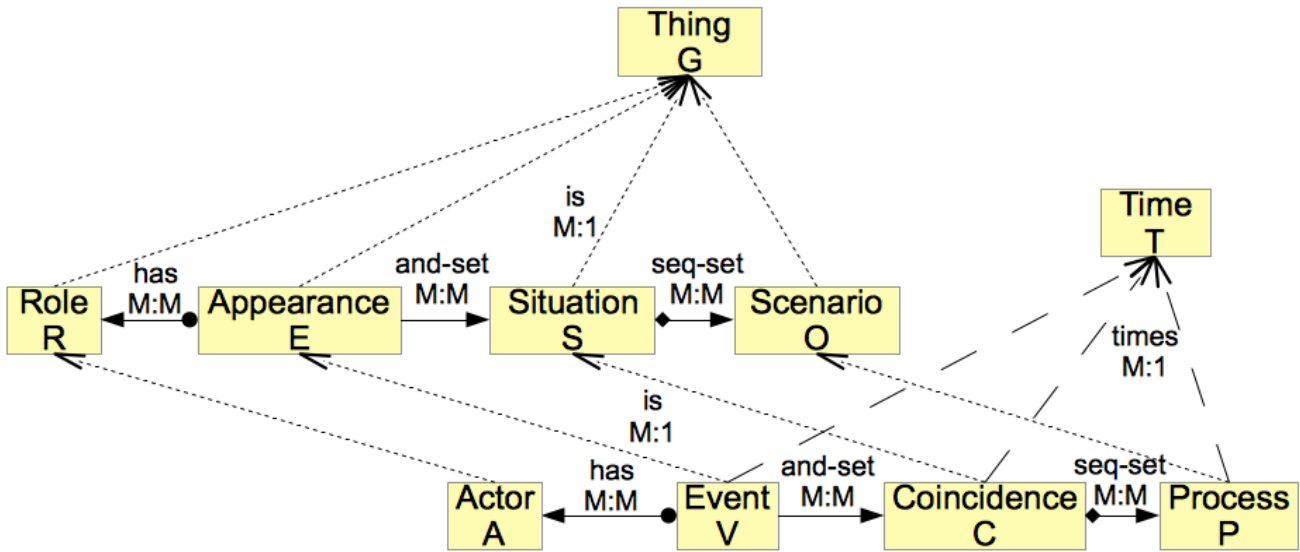

**Fig. 2.** Data model in use with types of relationships indicated.

Further, let us introduce required functional sets and functions in accordance with the definitions given above. Each of the functions may be considered as defining set in terms of boolean function, where **true** means inclusion in the set while **false** – exclusion from the set. It should be noted that such a boolean function may also be defined as probabilistic function in terms of fuzzy logic, so that the function may return value in range from 0 to 1 inclusively, and represent fuzzy set in such way. The following basic functional sets may be identified, with the possible extension that both input and arguments in the functions may be single things, probabilities of these things, sets, or sets with probability distributions:

$A(r)$ — all actors playing given role;
$R(a)$ — all roles of an actor;
$R(e)$ — all roles in given appearance, may be identified by collection of "**has**" links/edges in the graph;
$E(r)$ — all appearances involving given role, inverse to the previously defined one;
$E(v)$ — all appearances suitable to describe given event;
$E(t)$ — all appearances describing events corresponding to specified time moment or period, or probabilities of respective appearances at given of time;
$V(e)$ — all events representing given appearance;
$V(t)$ — all events at given moment or period of time or probabilities of that respective events may be taking place at given time, which can be called "event function" representing probability distribution of all possible events at a time;
$A(v)$ — all actors involved in an event in any role, $A(v,r)$ — all actors playing specific rules, identifying subgraph of links/edges of type r in larger unspecific graph;
$V(a)$ — all events involving particular actor, $V(a,r)$ — all events involving particular actor in specific role inversely to previously defined one, $V(a,r,t)$ — the above subgraph narrowed to the temporal slice specific to time t in subgraph specific to role type r;
$S(e)$ — all situations involving given appearance;
$E(s)$ — all appearance, involved in given situation;
$S(c)$ — all situations, generalizing given coincidence;
$C(s)$ — all coincidences, generalized by given situation;
$C(v)$ — all coincidences, including given event
$V(c)$ — all events, included in given coincidence;
$C(t)$ — all coincidences at given moment or period of time, $C(t,v)$ — including specific event;
$V(t)$ — all events at given moment or period of time, $V(t,c)$ — associated by specific coincidence;

O(s) — all scenarios, including given situation, O(s,q) — limited to sequential specified order;
S(o) — all situations involved in given scenario, S(o,q) — limited to sequential specified order;
P(t) — all processes taking place at given moment or period of time;
P(c) — all processes involving given coincidence, P(c,t) — limited to specific time moment or period;
O(p) — all scenarios that may be considered generalizing given process;
P(o) — all processes that me be considered generalized bu given scenario;
C(p) — all coincidences involved in given process, C(p,t) — limited to specific time moment or period;
T(a), T(v), T(c), T(p) — temporal continuums of existence of given actor, event, coincidence or process, respectively, all together identifying subgraph of links/edges of type "**times**".

It should be noted that functions like R(a), E(v), S(c) and O(p), based on respective scopes of definition, have their scopes of values identifying the set of relationships between specific (actors, events, coincidences, processes) and abstract (roles, appearances, situations and scenarios) things and may have respective reverse functions A(r), V(e), C(s) and P(o) applicable within the latter scopes.

For further discussion and related work it is important to bear in mind that notions of specific and abstract things may be thought as relative because the same thing may be specific in respect to more abstract thing in one context, while still being abstract in respect to more specific one in another context. For instance, the highly abstract statement "the person is doing something" can be thought of as an appearance with respect to the specific instance "this person is digging something" while the latter may be more abstract in respect to the even more specific "John is digging the ground" when it can still be further made more specific as "John Doe is digging the ground in the garden since noon". That means that all of the functional sets identified above may be thought not just as fuzzy sets, but they may also overlap with respect to the scopes of the function values.

The earlier warning in respect to the vagueness of the difference between appearance and the event should be kept in mind as well. That is, we keep the formal definition of event as having property of time (**times**) while appearances are thought to not be time-specific. However, this may be easily confused when one creates serval different appearances for "US president" for each of the actual US presidents and their terms, or if one takes the event of the sun rising on a given morning and keeps adding more and more mornings to the temporal scope of the event until the event of the sun rising becomes eternal phenomenon treated as appearance.

All the latter messages imply that any further practical developments would treat all definition dynamics from perspective of computational context, so that different things may play different roles in different stages of inferences, depending on the contextual scope and inference phase.

## 5. Unified algorithmic approach

In the following work we are going to consider the conceptual framework derived above applicable for implementation of any kind of cognitive functions — starting from recognition of written texts and spoken language and ending with causal analysis of high-level events identified in texts as well as in real life. That is, we consider the given approach to be applicable at multiple levels of generalization, so that the results of operations on lower abstraction levels can be aggregated and translated to higher abstraction levels where the same generic principles apply on each of the levels [6]. To justify this, we evaluate the applicability of our generic framework at different levels of abstraction and aggregation, identifying different sorts of things at every level, as shown in the Table 1 below.

|  | **Spoken language recognition** | **Written language recognition** | **Identification of patterns in the text** | **Causal analysis** |
|---|---|---|---|---|
| **Actor** | Coefficient on spectrogram for particular frequency | Property of specific stroke: On the top, at the bottom, long, short, skewed, etc. | Object property value: Name "John" | Specific actor: John Doe |
| **Event** | Combination of coefficients on spectrogram | Period or stroke composing the letter: . | Object class instance: Name "John", surname "Doe" | Specific event: John Doe cleaning the window on the second floor. |
| **Coincidence** | Specific sound | Coincidence: i | Co-occurrence of object of class person properties: "John Doe" | Specific coincidence: Window on the second floor is dirty, John Doe is cleaning it. |
| **Process** | Specific spoken word | Specific written word: ping | Specific phrase: "John Doe cleans the window" | Specific process: Window on the second floor was dirty, John Doe has cleaned it and now it is clean. |
| **Role** | Pitch frequency | Property of symbol: orientation, extent, symmetry, etc. | Domain of the object class property: person's surname | Typical role: Cleaner |
| **Appearance** | Spectral cluster on the spectrogram | Element of symbol: . | Class of the variable object: person | Typical appearance: Someone cleaning something |
| **Situation** | Sound of speech | Symbol: i | Pattern variable: $subject | Typical situation: Someone is cleaning something which is dirty. |
| **Scenario** | Spoken word accordingly to the language model | Written word: ping | Phrase pattern: "$subject cleans $object" | Typical scenario: Something was dirty, someone has cleaned it, it is clean now. |

**Table 1.** Correspondence of fragments of different cognitive functions to formal concepts of the suggested model — specific concepts (rows 1-4) and abstract concepts (rows 4-8).

Respectively, within suggested conceptual framework and mathematical model, we can identify different kinds of analysis which can allow to transfer cognitive activity between different abstraction levels with different degrees of data aggregation within cognitive functions of different kinds, as shown on Fig. 3.

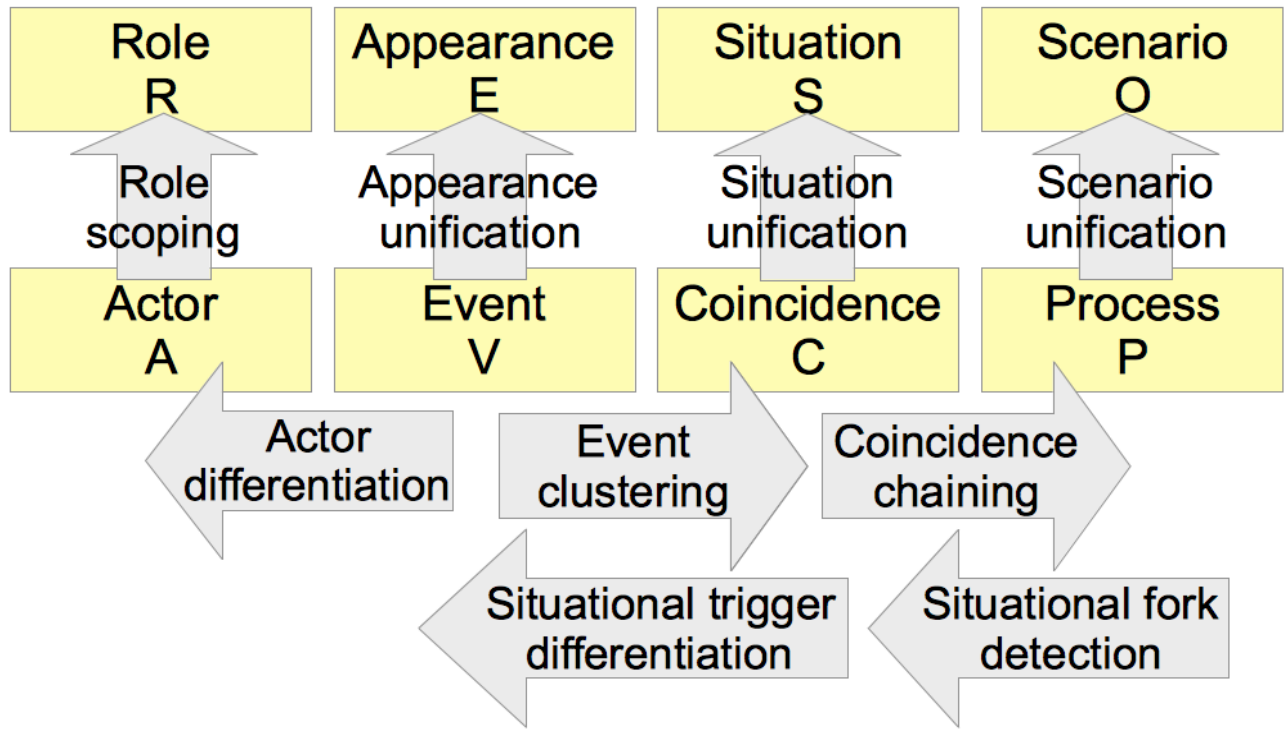

**Fig. 3.** Different kinds of analysis enabling to transfer focus of cognitive operations between low (at the bottom) and high (at the top) abstraction levels and different degrees of aggregation – atomic (on the left) and ordered (on the right).

- **Role scoping** — identification of the scope of actors providing the domain of possible valuations of a role based on multiple events instantiating the same appearance involving multiple alternative actors – from specific cases to general case. For example, the volume of known events related to stoplight operations identifies the single role "light color" with actors "red", "yellow" and "green" within abstract appearance of "stop light signal". In another example, specific events of cleaning the window by either mother or father identifies role of window cleaner with mother and father acting as actors in abstract appearance of window cleaning. Finally, events of perceiving the dot above an "i" and at the end of sentence and vertical stroke in letters "i" and "l" in different situations identify actors "dot" and "vertical stroke" as playing roles in different appearances of letters "i" and "l" as well as sentence termination.
- **Actor differentiation** — determination of individual actors playing similar roles in multiple events sharing other actors in different roles – from specific cases to the general case. For the first example above, position of the light on the stoplight panel can help differentiate different colors even for people suffering color blindness. For the second example, precedent analysis of multiple acts of window cleaning evidence may identify the mother as the most probable candidate for playing this role in general case. For the last example, multiple appearances of "i" and "." with dot residing in different positions in different neighborhoods may suggest different roles played by the actor in different appearances – as either topping of "i" or self-sufficient symbol.
- **Appearance unification** — identification of sets of specific events being instances of the same generic abstract appearance, even if actors in different roles are changing. That is, the abstract operation of the stoplight signal may have different actors in roles "light color" and "light position", still being part of the "stoplight signal" appearance. Respectively, appearance of window cleaning may become generalized so no actor playing the role of the cleaner matters, be it the mother, the father or the home service person. Lastly, no matter if a dot is placed far above the vertical stroke or close to it, it will be identified as part of "i", at the same time that it isbreaking the sentence, as long as it is placed at the bottom of the text line.

- **Event clustering** — determination of sets of coinciding events sharing the same time property as related to certain group of appearances. For instance, "red" actor playing the "light color" role in "stoplight signal" appearance may be associated with coinciding car movements across the crosswalk – the coincidence easily learned by dogs and geese. Similarly, the event of window cleaning may be coinciding with the inability to play toys on the window-sill while a dot appearing at the end of the text line may be found to coincide with a subsequent capital letter starting the next sentence.
- **Situation unification** — identification of generalities across different coincidences sharing similar combinations of events belonging to different groups, i.e. representing different typical situations. For one case, typical situations may be a combination of "green light" and "cars on the crosswalk", for another case, it could be a combination "window is being cleaned", "parent is busy" and "no games nearby", in the third case it could be "dot at the bottom" with "next letter is capital" as a sentence break.
- **Coincidence chaining** — determination of chains of events following one after another so the continuous and logically connected process is built up. One example could be waiting at green and flashing yellow lights, then safe crossing the crosswalk on green light. The other example could be mother or father coming into the room with some water and rag, then cleaning the window and then leaving the window with more sun light coming into the room from outside. The third example could be a text tokenization process reaching the dot at the very end of non-numeric token, then finding single whitespace character followed by any capital letter, so the decision to break the sentence could be made.
- **Scenario unification** — identification of statistically reliable chains of temporally adjacent coincidences with the same order of situations that they are instantiating, in other words – finding the set of identical or similar enough processes residing on different intervals of a temporal axis. The scenarios could be "safe crossing the crosswalk on the green light", "injury when crossing the crosswalk on the red light", "Saturday window cleaning in a children's room" or "sentences breaking on a dot, closing the non-numeric token, with a subsequent space and capital latter heading the subsequent token".
- **Situational fork detection** — this process may be applied to a family of scenarios, sharing the same initial chain, i.e. sharing identical or similar sequences of situations until some point, but having different situations past this point, when the base scenario forks into alternative ones. For one instance, entering the crosswalk by pedestrians while the red light is on may cause alternative scenarios involving sequences of events such as emergency braking, car skidding and hurting the bunch of people waiting for green light. In another instance, the window cleaning scenario may be altered by throwing the car toy into the window having the latter smashed so there is not anything to clean from this point until repair works are accomplished. Lastly, a sentence break identification scenario based on dots at the low level are treated as periods that may be interrupted with a numeric character following the period itself so that it could be treated as part of a web site or file name. This kind of analysis, Markov hidden chains may be involved.
- **Situational trigger differentiation** — for very practical purposes, it may be possible to identify specific appearances or situations, so that the addition of their instances as events or coincidences to the process in the course of development may change the effective scenario from one to another. That is, for the family of scenarios with shared initial segments and situational forks identified, it could be possible to identify triggering appearances or situations which makes the scenarios different, causing alternative outcomes. In the first example, entering the crosswalk on red light is one such trigger, causing health damage and possible death in the end. In the second case, the toy car thrown into the window is a cause preventing the window cleaning procedure but involving other unpleasant consequences. In the second case, omission of whitespace past the period may trigger missing sentence break and false identification of web site name in the text being tokenized.

## 6. Conclusion

The conceptual framework above does not address the actual mechanisms of the inferences implementing the inferential processes discussed in the previous chapter. We anticipate that any of probabilistic or fuzzy logic frameworks [7,8,9] may be used for this as well as novel suggestions. Among the latter ones, one promising approach would be to explore the applicability of "disjuncts" (as events and appearances) and "connectors" (as roles) suggested for the unsupervised learning of linguistic structures [10] to be applied to a wider scope of domains.

## References

1. S. S. Nandagaonkar, D.B.Hanchate, S.N. Deshmukh: Survey on Event tracking and Event Evolution, Int.J.Comp.Tech.Appl,Vol 3 (1), 1-4, ISSN:2229-6093, 2012
http://www.ijcta.com/documents/volumes/vol3issue1/ijcta2012030101.pdf
2. Anton Kolonin: Intelligent Agent for Web Watching : Language and Belief System, Problems of Informatics, ISSN 2073-0667, Issue 2, 2015, pp.59-69.
http://aigents.com/papers/2014AgentBeliefKolonin.pdf
3. Anton Kolonin: Automatic text classification and property extraction, 2015 SIBIRCON/SibMedInfo Conference Proceedings, ISBN 987-1-4673-9109-2, pp.27-31
https://ieeexplore.ieee.org/document/7361868/?arnumber=7361868
4. Kazutoshi Sumiya, Minori Takahashi, Katsumi Tanaka: WebSkimming : An Automatic Navigation Method along Context-Path for Web Documents, The Eleventh International World Wide Web Conference, Journal Article, 2002
http://wwwconference.org/proceedings/www2002/poster/202/index.html
5. Yingqin Gu, Jianwei Cui, Hongyan Liu, Zhixu Li: Detecting Hot Events from Web Search Logs, Web-Age Information Management: 11th International Conference, WAIM 2010, Jiuzhaigou, China, July 15-17, 2010. Proceedings (pp.417-428)
https://www.researchgate.net/publication/221509606_Detecting_Hot_Events_from_Web_Search_Logs
6. Gerhard Werner: Fractals in the Nervous System: Conceptual implications for Theoretical Neuroscience. Front Physiol. 2010; 1: 15. Published online 2010 Jul 6. Prepublished online 2010 May 26. doi: 10.3389/fphys.2010.00015
https://www.ncbi.nlm.nih.gov/pmc/articles/PMC3059969/
7. Pei Wang: Non-Axiomatic Reasoning System (Version 4.1)
The Seventeenth National Conference on Artificial Intelligence, 1135-1136, Austin, Texas, July 2000
https://cis.temple.edu/~pwang/Publication/NARS-41.pdf
8. E.E. Vityaev, L.I. Perlovsky, B.Y. Kovalerchuk, S.O. Speransky: Probabilistic dynamic logic of cognition. Biologically Inspired Cognitive Architectures. Special issue: Papers from the Fourth Annual Meeting of the BICA Society (BICA 2013), v.6, October, Elsevier, 2013, pp.159-168.
http://www.math.nsc.ru/AP/ScientificDiscovery/PDF/probabilistic_dynamic_logic_of_cognition_bica.pdf
9. Ben Goertzel, Matthew Iklé, Izabela Freire Goertzel, Ari Heljakka: Probabilistic Logic Networks: A Comprehensive Framework for Uncertain Inference, 1st Edition. 2nd Printing. 2008 Edition, Springer, 2008, ISBN-13: 978-0387768717, ISBN-10: 0387768718
https://www.amazon.com/Probabilistic-Logic-Networks-Comprehensive-Framework/dp/0387768718
10. Linas Vepstas, Ben Goertzel: Learning Language from a Large (Unannotated) Corpus,

https://arxiv.org/abs/1401.3372